\documentclass[runningheads]{llncs}
\usepackage{graphicx}
\begin{document}
\title{IITP at AILA 2019: System Report for Artificial Intelligence for Legal Assistance Shared Task}
 \author{Baban Gain \inst{1}\and
 Dibyanayan Bandyopadhyay\inst{1} \and
 Arkadipta De\inst{1}\and
 Tanik Saikh\inst{2}\and
 Asif Ekbal\inst{2}}
% %
 \authorrunning{B. Gain et al.}
 \titlerunning{IITP at AILA 2019: System Report for AILA Shared Task }
 \institute{Government College Of Engineering And Textile Technology, Berhampore \and
 Indian Institute of Technology Patna  \\
 \email{\{gainbaban,dibyanayan,de.arkadipta05\}@gmail.com\inst{1}}\\
 \email{\{tanik4u,asif.ekbal\}@gmail.com}\inst{2}}
\maketitle              % typeset the header of the contribution
\begin{abstract}
In this article, we present a description of our systems as a part of our participation in the shared task namely \textit{Artificial Intelligence for Legal Assistance (AILA 2019)}. This is an integral event of \textit{Forum for Information Retrieval Evaluation - 2019}. The outcomes of this track would be helpful for the automation of the working process of the Indian Judiciary System. The manual working procedures and documentation at any level (from lower to higher court) of the judiciary system are very complex in nature. The systems produced as a part of this track would assist the law practitioners'. It would be helpful for common men too. This kind of track also opens the path of research of Natural Language Processing (NLP) in the judicial domain. This track defined two problems such as \textit{Task 1 : Identifying relevant prior cases for a given situation and Task 2 : Identifying the most relevant statutes for a given situation}. We tackled both of them. Our proposed approaches are based on \textit{BM25} and \textit{Doc2Vec}. As per the results declared by the task organizers', we are in 3rd and a modest position in Task 1 and Task 2 respectively.

\end{abstract}

\section{Introduction}
The working procedure of any judiciary system over the world is very complex in nature. \footnote{Copyright © 2019 for this paper by its authors. Use permitted under Creative Commons License Attribution 4.0 International (CC BY 4.0). FIRE 2019, 12-15 December 2019, Kolkata, India.} The countries like India are having two primary sources of law namely: \textit{Statutes (established laws)} and \textit{Precedents (prior cases)}. Statutes are applied legal principles to a situation like facts/scenario/circumstances which lead to filing the case. Whereas, the precedents/prior cases help a lawyer to understand how the court has dealt with similar scenarios in the past, and make the reasoning accordingly. These kinds of tasks are very cumbersome and hard to do manually. Automated solutions to these problems would be beneficial to the lawyers. The system will do the task of finding relevant prior cases given a current case and statutes/acts that will be more suited to a given situation. These kinds of systems will be helpful to common men too. It will be able to provide a preliminary understanding to common men on a particular case even before going to the lawyers. It shall assist human beings in identifying where his/her legal problem fits, what legal actions he/she can proceed with (through statutes), and what were the outcomes of similar cases (through precedents). The \textit{Artificial Intelligence for Legal Assistance
(AILA 2019)} \cite{fire2019-aila}, an associated event of  \textit{Forum for Information Retrieval Evaluation 2019} has come up with two problems \textit{viz. i. Task 1 : Identifying relevant prior cases for a given situation and ii. Task 2 : Identifying the most relevant statutes for a given situation}. The automated solutions to these problems will mitigate such problems of lawyers as well as common men. We took part in both the tasks defined. We make use of the datasets released for these tasks for our experiments. The dataset is having a set of 50 queries, each of which contains the description of a situation. In Task1, almost 3000 case documents of cases that were judged in the Supreme Court of India are given. The participants of this task have to retrieve the most similar/ relevant case documents with respect to the situation in the given query. The dataset for the Task 2 is having 197 statutes (Sections of Acts) from Indian law, that are relevant to some of the queries. The title and the description of these statutes are given. Task 2 is to identify the most relevant statutes (from among the 197 statutes) for each query. These tasks could be tackled either in supervised or unsupervised way. We make use of unsupervised approaches as the datasets are having very less number of annotated examples.
\section{Related Work}
Earlier this type of work was introduced by \cite{DBLP:conf/fire/MandalGBPG17}. They defined two tasks namely \textit{i. Catch Phrase Extraction and ii. Precedence Retrieval}. This track received a lot of responses. Another task namely \textit{Competition on Legal Information Extraction and Entailment (COLIEE-2019)}, as an associated event of \textit{International Conference on Artificial Intelligence and Law (ICAIL)-2019} was performed in the same line. In this competition, four tasks were defined. Task 1 was to retrieve the supporting cases for a new case from the whole case law corpus. Task 2 was about to identify paragraphs from prior cases that entails the decision of a new case. Task 3 was a legal question answering task. Task 4 was to retrieve relevant documents given a question and then have to determine whether the relevant articles entail the question or not. The work of \cite{Tran:2019:BLC:3322640.3326740} tackled all the problems. They made use of BERT, BM25 and Doc2Vec models for these problems. 
\section{Proposed Methods and Experimentation}
We participated in both the tasks defined. The first task is to identify relevant prior cases for a given situation and the other one is identifying the most relevant statutes for a given situation. They provided datasets for both the tasks. The dataset contains 50 queries. The queries are a description of legal situations. For Task 1, there are 2,914 prior case documents and for Task 2, 197 statutes are there which are basically the title and the corresponding textual description of the statutes. We foster BM25 \cite{robertson2009probabilistic} and Doc2Vec \cite{Le:2014:DRS:3044805.3045025} approaches for the Task 1. Only BM25 has been used for Task 2. 
\subsection{Task 1}
In this part, we discuss the preprocessing of the datasets for this task followed by the description of the proposed approaches and experimental procedures.\\ 
\textbf{\textit{Preprocessing:}} We perform a few preprocessing first. For every query and candidate document, we extract all the words. Then, remove all the numbers, that are used to indicate paragraph number. We also remove the stop words from the documents by using NLTK English stop words list. Then we perform stemming and lemmatization of the words using Porter Stemmer and WordNet Lemmatizer respectively. \\
\textbf{\textit{BM25:}}
BM25 is a bag-of-words based retrieval function that ranks a set of documents based on the query terms appearing in each document, regardless of the inter-relationship between the query terms within a document. The objective of the task is to retrieve the most similar/relevant case documents concerning the situation of a given query document. The dataset contains 2914 candidates and 50 queries, among which 10 queries are used as train cases and 40 queries are used for the testing purpose as directed by the task organizers'. \\ 
\textbf{\textit{Doc2Vec:}} Doc2Vec is an unsupervised algorithm that learns fixed-length feature representations from variable-length pieces of texts, such as sentences, paragraphs, and documents. This algorithm represents each document by a dense vector that is trained to predict words in the document. More precisely, we concatenate the paragraph vector with several word vectors from a paragraph and predict the following word for the given context. Both word vectors and paragraph vectors are trained by the stochastic gradient descent and backpropagation \cite{rumelhart1988learning}. While paragraph vectors are unique among paragraphs, the word vectors are shared. At prediction time, the paragraph vectors are inferred by fixing the word vectors and training the new paragraph vector until convergence. \\
We submitted two runs for the task:
\begin{itemize}
  \item \textit{$IITP\_BM25\_case$}: We compute the BM25 similarity score (as mentioned earlier) for a query document with every candidate cases, and then the candidates are returned as in decreasing order of the similarity score between a query document and a candidate case.

  \item \textit{$IITP\_Doc2Vec\_case$}: We train a gensim Doc2Vec model using candidate cases and query documents. The hyperparameters applied are as follows: Vector dimension: 150, window: 20, epoch: 50. After training, we get document vectors for every query and candidate document. Then we calculate the Doc2Vec similarity score between the query document and every candidate document. Then the candidates are returned in decreasing order of similarity score with the query.

\end{itemize}

\subsection{Task 2}
The objective of this task is to identify the most relevant statues for a given query document. The dataset consisting of 197 statutes and 50 queries among which 10 queries are used for training purpose and 40 queries for testing purpose. \\
\textbf{\textit{Preprocessing:}} For every query and statute, we extract all the words. We perform removal of stop words, stemming and lemmatization of words using NLTK English stop words list, Porter Stemmer and WordNet Lemmatizer respectively.\\
We submitted one run for this task:\\
\textbf{\textit{$IITP\_BM25\_statute$}}:
This is our only approach to this task. We compute the BM25 similarity score between a query document and every statute and then the candidates are returned in decreasing order of similarity score. \\

\section{Results and Discussion}
This section describes the results obtained in the two tasks by our proposed approaches. The results obtained in Task 1 are shown in the Table \ref{resu-task1}. In the Table, MAP is mean average precision, BPREF is a measure used when the relevance assessments are not enough and it is suspected that there are many documents (usually the relevant ones) which are not considered for assessment. In this Table, the last column i.e. column 5 indicates \textit{1 / rank of the first relevant document} as provided by the task organizers'. It is basically, \textit{Mean Reciprocal Rank} used for exact item search where we are interested in finding *one* (the first one) correct answer and not *all* correct ones.
\begin{table}[]
\centering
\caption{Results obtained in Task 1 using two methods. }
\label{resu-task1}
\begin{tabular}{|l|l|l|l|l|}
\hline
\textbf{Run ID} & \textbf{Prec@10} & \textbf{MAP} & \textbf{BPREF} & \textbf{Rank} \\ \hline
\textbf{IITP\_BM25\_case} & 0.0275 & 0.0984 & 0.0869 & 0.175 \\ \hline
\textbf{IITP\_Doc2vec\_case} & 0.0175 & 0.0677 & 0.0552 & 0.138 \\ \hline
\end{tabular}
\end{table}
The first run, i.e. \textbf{IITP\_BM25\_case} model achieved a MAP of 0.0984 and BPREF of 0.0869, which are the 5th among all 23 runs submitted for Task 1 by various teams. In terms of runs submitted by unique team, this run is in 3rd position in terms of MAP and BFREF. \\ 
The results obtained in Task 2 are shown in Table \ref{resu-task2}.  
\begin{table}[]
\centering
\caption{Results obtained in Task 2 using the proposed approach. }
\label{resu-task2}
\begin{tabular}{|l|l|l|l|l|}
\hline
\textbf{Run ID} & \textbf{Prec@10} & \textbf{MAP} & \textbf{BPREF} & \textbf{rank} \\ \hline
\textbf{IITP\_BM25\_statutes} & 0.02 & 0.036 & 0.0397 & 0.129 \\ \hline
\end{tabular}
\end{table}
In this task (i.e. Task 2), we obtain a secure position in the leaderboard. 

\section{Conclusion and Future Work}
This paper presents the working note of our participation in \textit{Artificial Intelligence for Legal Assistance
(AILA 2019)} track, which is an associated event of \textit{Forum for Information Retrieval Evaluation (FIRE)- 2019}. The shared task defined two tasks as follows: \textit{viz. i. Identifying relevant prior cases for a given situation ii. Identifying the most relevant statutes for a given situation}. We participated in both the tasks. We submitted two runs for task1 and one for the Task 2. In Task 1, our models are based on Doc2Vec and BM25. We proposed the BM25 based model for Task 2. We stood 3rd in Task 1 and secured a moderate position in the scoreboard of the second task. Our future work including:
\begin{itemize}
   \item Development of deep learning based supervised approach. As this kind of approach itself is very data hungry, we need to increase the volume of the data.  
   \item We could make use of several other unsupervised similarity metrics for these tasks.
   \item Contextual embedding followed by attention mechanism might be a good approach to tackle these tasks. But for all the cases data is the main bottleneck. We should put our focus on preparing more judicial domain annotated data for these problems.
   \item We could enlarge this dataset by merging the dataset which was released in \textit{Information Retrieval from Legal Documents (IRLeD) - FIRE 2017}. That dataset is in the same line of this track i.e. \textit{Artificial Intelligence for Legal Assistance (AILA) 2019} dataset.   
\end{itemize}
\section{Acknowledgments}
Mr. Tanik Saikh acknowledges all the co-authors for their contributions. We also acknowledge the funding agency for the commitment to providing funds.

\bibliographystyle{splncs04}
\bibliography{mybibliography}




% LNCS Homepage, \url{http://www.springer.com/lncs}. Last accessed 4
% \end{thebibliography}
\end{document}